\documentclass[preprints,article,accept,moreauthors,pdftex]{Definitions/mdpi}

\newcommand{\ie}{\textit{i}.\textit{e}.}

\newcommand{\wrt}{\textit{w}.\textit{r}.\textit{t}.}

\usepackage{cleveref}
\usepackage{subfig}
\usepackage{graphicx}
\usepackage{gensymb}

\firstpage{1} 
\pubvolume{xx}
\issuenum{1}
\articlenumber{5}
\pubyear{2019}
\copyrightyear{2019}
\history{Received: date; Accepted: date; Published: date}

\Title{A Termination Criterion for Probabilistic Point Clouds Registration}

\Author{Simone Fontana $^{1}$ and Domenico G. Sorrenti$^{1}$}

\AuthorNames{Simone Fontana and Domenico G. Sorrenti}

\address{%
$^{1}$ \quad Università degli Studi di Milano - Bicocca ; name.surname@unimib.it}

\corres{Correspondence: simone.fontana@unimib.it}

\abstract{Probabilistic Point Clouds Registration (PPCR) is an algorithm that, in its multi-iteration version, outperformed state of the art algorithms for local point clouds registration. However its performances have been tested using a fixed high number of iterations. To be of practical usefulness, we think that the algorithm should decide by itself when to stop, to avoid an excessive number of iterations and, therefore, wasting computational time. With this work, we compare different termination criterion on several datasets and prove that the chosen one produce very good results that are comparable to those obtained using a very high number of iterations, while saving computational time.}


\begin{document}


\section{Introduction}
Point clouds registration is the problem of finding the transformation (mostly a rigid transformation) that best aligns two point clouds, usually called the \emph{source} and \emph{target} point clouds.

One of the first approaches to this problem, and still one of the most used, is Iterative Closest Point (ICP) \cite{besl1992method, chen1991object,zhang1994iterative}, which aligns two point clouds by minimising the sum of distances between corresponding points, where corresponding points are nearest neighbouring points. Probabilistic Point Clouds Registration (PPCR) \cite{agamennoni2016point} is variant of ICP that uses a probabilistic model to improve the robustness against noise and outliers, one of the most relevant problem of local registration algorithms. Much like ICP, it is an iterative algorithm that repeatedly tries to improve a solution, until a convergence criterion is satisfied. 

The experiments show that it outperformed most state-of-the-art local registration algorithms in this field. However, these experiments have been performed using a large fixed number of iterations as stopping criterion. Instead, we think that, to be of practical utility, an iterative algorithm should autonomously decide when to stop. Indeed, using a fixed number of iterations on one hand does not guarantee that the best solution have been found; on the other, it could result in an excess of computation time, because the solution have been found earlier. 

For these reasons, we propose an improvement of PPCR, analysing different termination criteria and finding the best one. Moreover, we demonstrate that the chosen solution is as effective as using a very high number of iterations, but, at the same time, results in fewer iterations and, therefore, less computational time.

\section{Related Work}
\label{sec:related_work}
Point clouds registration algorithms could be divided into two categories: local and global. Local registration algorithms (also known as \emph{fine} registration) aim at finding the rototranslation that best aligns two point cloud that are already roughly aligned. Therefore, they refine a pre-existing alignment, that can be obtained in different ways; examples are: with another algorithm, with an inertial system or manually.

One of the most important algorithms in this category is Iterative Closest Point (ICP).  ICP was developed independently by Besl and McKay \cite{besl1992method}, Chen and Medioni \cite{chen1991object}, and Zhang \cite{zhang1994iterative} and is still one of the most used technique. The most critical problem a registration algorithm has to solve is the data association problem, that is, associating one point in a point clouds, to one or more in the another. ICP solves this issue by associating to a point in the source point clouds the closest in the target. The best transformation resulting from this data association is found and this process is repeated until convergence. 

Many different variants of ICP have been proposed. Usually, they aim at speeding up the algorithm or at improving the accuracy \cite{pomerleau_comparing_2013}. One of the most important of these variants is Generalized ICP (G-ICP) \cite{segal_generalized-icp._2009}, that greatly improves the quality of the results by using a probabilistic framework with a point-to-plane data association.

Probabilistic Point Clouds Registration (PPCR) \cite{agamennoni2016point} uses the same closest-point based data association of ICP, in conjunction with a probabilistic model, to improve both the accuracy and, most important, the robustness against noise and outliers. While it was originally developed to deal with the problem of aligning a sparse point cloud with a dense one, it was shown to perform very well also on traditional registration problems.

Another important technique used for local point clouds registration is called Normal Distribution Transform (NDT),~\cite{biber2003normal}. This technique was originally developed to register 2D laser scans, but has been successfully applied also to 3D point clouds,~\cite{merten2008three}. Differently from ICP, it does not establish any explicit correspondence between points. Instead, the source point cloud or laser scan is subdivided into cells and a normal distribution is assigned to each cell, so that the points are represented by a probability distribution. The matching problem is then solved as a maximization problem, using Newton's algorithm.

The second category of point clouds registration algorithms aims at global registration, that is, aligning two point clouds without any prior assumption on their misplacement. Traditionally, this problem has been solved using features-based techniques, such as PFH~\cite{rusu_aligning_2008} and their faster variant FPFH~\cite{rusu_fast_2009}, or angular-invariant features,~\cite{jiang_registration_2009}. Usually the matches found are used to estimate the rototranslation between the two point clouds using algorithms such as RANSAC \cite{fischler1981random}. As an alternative to \emph{hand-crafted} descriptors, solutions based on neural networks, that aim at enhancing the discriminative capacity of the features, have been proposed. Examples are 3dMatch \cite{zeng20173dmatch} and 3DSmoothNet \cite{gojcic2019perfect}. Networks that combines both the feature matching and the transformation estimation steps together have been proposed too, such as Pointnetlk \cite{aoki2019pointnetlk} and Pcrnet \cite{sarode2019pcrnet}.

The drawback of global registration approaches is that they usually cannot provide an accurate alignment, mainly because of the high number of spurious matches; therefore, they are rather used to obtain a coarse registration that is later refined with a fine registration algorithm ~\cite{holz2015registration}. For this reason, techniques aimed at estimating a rototranslation from matches with a high number of outliers have been proposed. Notable examples are Fast Global Registration \cite{zhou2016fast} and TEASER++ \cite{Yang20arXiv-TEASER}, that can even work without any feature, but using an all-to-all association strategy. While these approaches are a great improvements over traditional feature-based techniques, they have not been proved yet to outperform the best local registration algorithms.

\section{Materials and Methods}
\subsection{Probabilistic Point Clouds Registration}
We already presented PPCR in a previous work \cite{agamennoni2016point}; however, since we present an extension to the original version, we briefly summarize its working.

PPCR is a closest-point based algorithm for local point clouds registration. This means that it is aimed at fine-aligning two point clouds that are already roughly aligned. It does not use any feature to estimate correspondences between two point clouds; instead, similarly to ICP, it approximates the true, unknown, correspondences by using a data-association policy based on the closest distance.

However, our data association policy differs from that of ICP (and many of its variants): in ICP each point in the source point cloud is associated only with a point in the target point cloud, while PPCR associates a point in the source point cloud with a set of points in the target cloud. Moreover, the associations are weighted. The weights represent the probability of an association of being the right data-association for a particular point.

\begin{figure}
\centerline{
\subfloat[]{\includegraphics[width=0.2\linewidth]{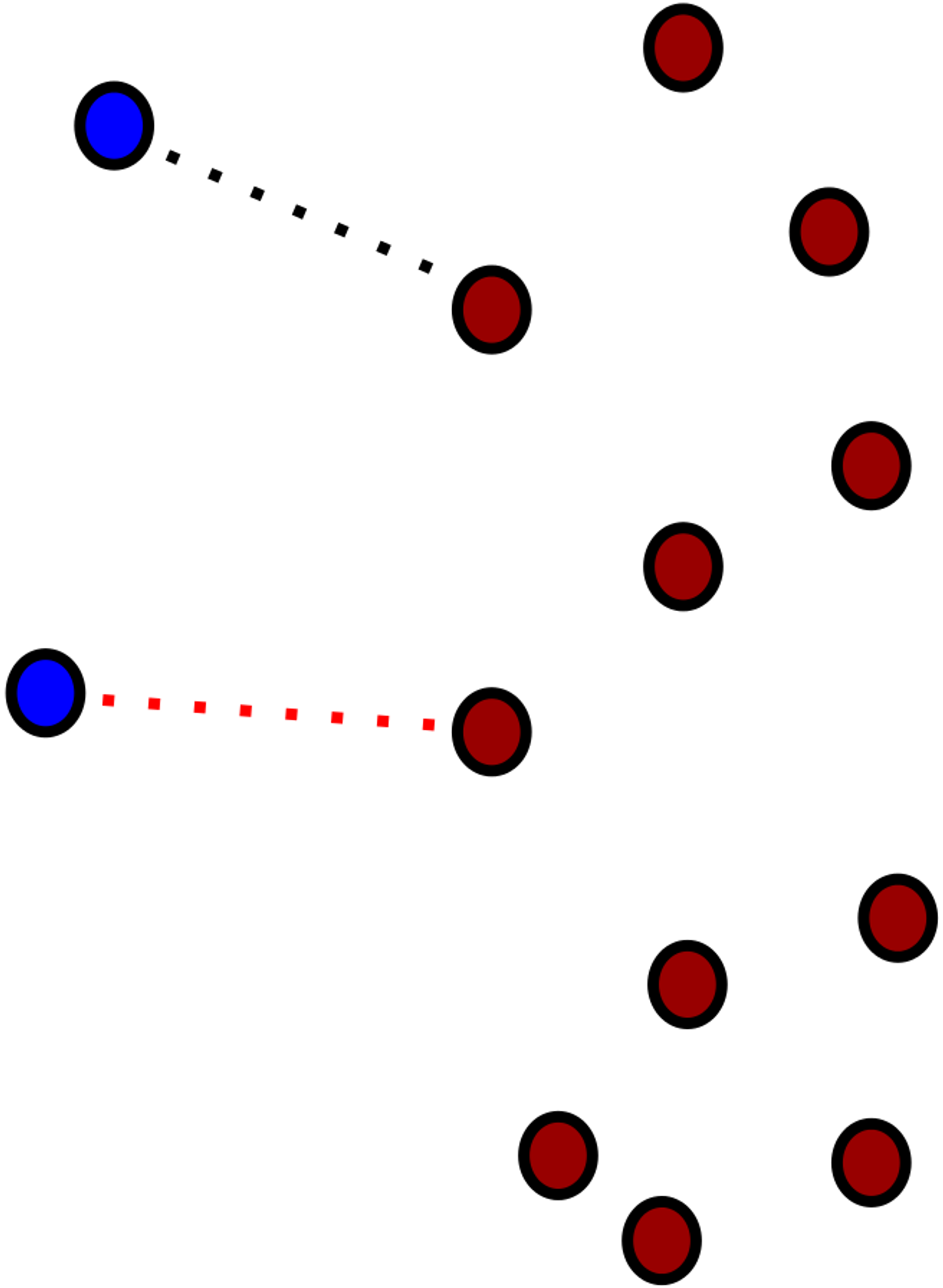}
\label{fig:icpDataAssoc}}
\hfil
\subfloat[]{\includegraphics[width=0.2\linewidth]{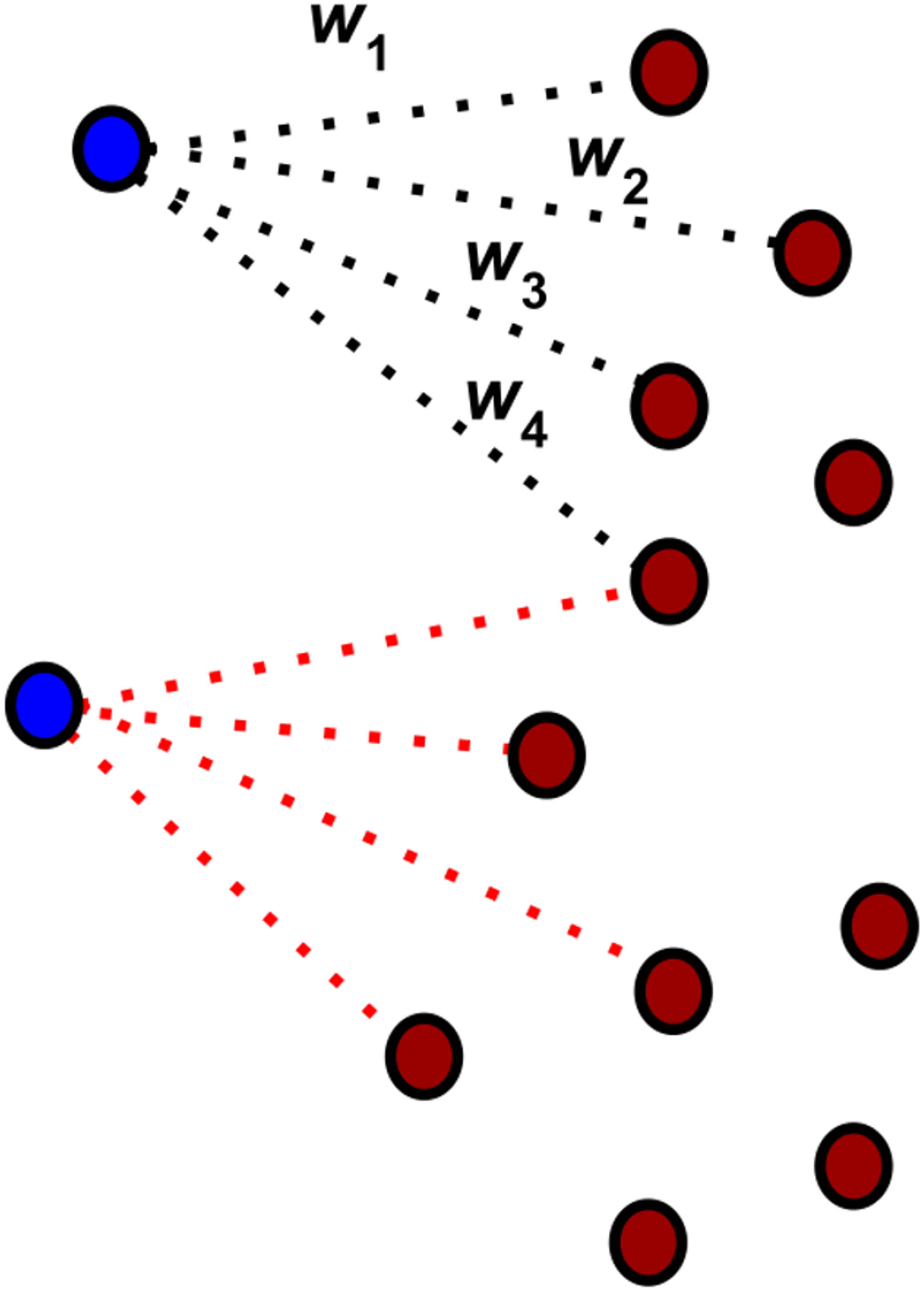}
\label{fig:probDataAssoc}}}
\caption{The two different data association policies. (a) ICP Data Association (b) Probabilistic Data Association}
\label{fig:dataAssoc}
\end{figure}

The two different data association methods are depicted in \cref{fig:dataAssoc}

For each point $x_j$ in the source point cloud, we look for the $n$ nearest points, $y_0, ..., y_n$, in the target cloud. For each of these points $y_k$, with $0\leq k \leq n$, we define an error term given by
\begin{equation}
\label{eq:squared_error}
\| y_k - (Rx_j + T) \|^2
\end{equation}

Equation~\ref{eq:squared_error} represents the squared error between the point $y_k$ in the target point cloud and the associated point $x_j$ from the source point cloud, transformed using the current estimate of the rototranslation. 

The set of error terms, calculated according to Equation \ref{eq:squared_error}, forms an optimization problem which is solved using a suitable method (such as Levenberg-Marquardt). However, given a set of points associated to $x_j$, not all the corresponding error terms should have the same weight. Intuitively we want to give more importance to the associations that are in accordance with the current estimate of the transformation and lower importance to the others. Thus, the weight of the error term $\| y_k - (Rx_j + T) \|^2$ is given by
\begin{equation}
\label{eq:gaussian_weight}
w_{kj} \propto e^{-\frac{\| y_k - (Rx_j + T) \|^2}{2}}
\end{equation}
where the proportionality implies a normalization among all the error terms associated with $x_j$ so that their weights represents a probability distribution. Equation \ref{eq:gaussian_weight} is derived from the EM algorithm, with an additive Gaussian noise model \cite{agamennoni2016point}.

The Gaussian in Equation~\ref{eq:gaussian_weight} is appropriate assuming that there are no outliers and all points in the source point cloud have a corresponding point in the target point cloud. However, a t-distribution is a better choice in presence of outliers, especially when there is lot of distortion in one of the point clouds that, thus, cannot be aligned perfectly. Consequently, we decided to use a more robust formulation for the weights, basing on the t-distribution. A t-distribution is very similar to a Gaussian, but the its tails have a higher probability; therefore, the probability of having outliers is higher than when using a Gaussian.
\begin{equation}
\label{eq:t_prob}
p_{kj} \propto \left(1+\frac{\| y_k - (Rx_j + T) \|^2}{\nu} \right)^{-\frac{\nu + d}{2}}
\end{equation}
\begin{equation}
\label{eq:t_weight}
w_{kj} = p_{kj}\frac{\nu+d}{\nu +\| y_k - (Rx_j + T) \|^2 }
\end{equation}
where $\nu$ the is number of degrees of freedom of the t-distribution and $d$ is the dimension of the error terms (in our case 3, since we are operating with points in the 3D space).

In \cref{eq:t_prob} we need an estimate of the rotation and translation; however these are estimated by solving the optimization problem whose error terms are weighted with the weights we want to calculate. Hence our problem cannot be formulated as a simple least-square error problem, but it has to be reformulated as an Expectation Maximization problem. During the Expectation phase the latent variables, in our case the weights, are estimated using the previous iteration estimate of the target variables (the rotation and translation), while during the Maximization phase, the problem becomes a least-square error optimization problem, with the latent variables assuming the values estimated during the Expectation phase.

The proposed approach, in its multi-iteration version, is composed of two nested loops. The inner one finds the best rototranslation that minimizes the sum of weighted squared errors (as in \cref{eq:squared_error}, very similarly to ICP. However, differently to ICP, our problem cannot be solved in closed form and, thus, we use an iterative algorithm such as Levenberg-Marquard. Notice that at each iteration of Levenberg Marquard, the associations are not estimated again, but their weights are recalculated. Thus we solve an iteratively reweighted mean squared error problem.
 In the outer loop, we move the source cloud with the result of the optimization, re-estimate the associations and build a new optimization problem. This structure has been already briefly described in our previous work. However here we present a novel way to decide when the outer loop should stop, instead than using a predefined number of iterations.

\subsection{Termination Criteria}
In case that the source and the target point clouds are very close, a single iteration of the proposed probabilistic point clouds registration algorithm may be enough. However, in a typical real scenario, more iterations are necessary.
In order for the algorithm to converge, most of the correspondences used to form the optimization problem need to be right. Since we use a data association policy based on the euclidean distance,  this happens only if the two point clouds are close enough. 
In our algorithm, two parameters control which and how many points in the target point cloud are associated to a particular point in the source point cloud: the maximum distance between neighbors and the maximum number of neighbors. Setting these parameters to very high values could help the algorithm to converge to a good solution even when the starting poses of the two point clouds are not really close. However, this will allow more outliers, \ie{}, wrong data associations, to get into the optimization step. Even tough the probabilistic approach has the capability to soft-filtering out outliers, thanks to the probabilistic weighting technique, using too many points will lead to a huge optimization problem which would be very slow to solve. Usually, a much more practical and fast solution is to use lower values for the maximum distance and the maximum number of neighbors and use multiple iterations of the probabilistic approach, that implies re-estimating the data associations, in the same way it is done, for example, in ICP and G-ICP.

With this technique, our approach becomes composed of two nested loops. The inner one solves an optimization problem using  the Levenberg-Marquard algorithm. The outer one moves the point cloud using the solution found in the inner loop, re-estimates the point correspondences and builds the corresponding optimization problem. This process is repeated until some convergence criterion is met.

The multi iteration version of our algorithm provides good results, compared to other state of the art algorithms \cite{agamennoni2016point}. Of course, in order to be of practical usefulness, such an algorithm would greatly benefit from some kind of automatic termination criterion. It would mean that the algorithm could decide by itself when it should stop.

The most simple termination criterion is to use a fixed predefined number of iterations. This is the technique we used in our previous work. However, this solution is far from being optimal, since the number of iterations would become a parameter of the algorithm. Most importantly, there would be no automatic way of estimating this parameter \emph{a-priori}, so this solution is unpractical and has to be discarded. Lastly, using a fixed value for this parameter would probably mean using too many iterations in some cases and using too few in others. On the other hand, using a very large value would greatly increase the execution time, in many cases without improving the quality of the result.

For these reasons, we evaluated different automatic termination criteria, to find which one works best with PPCR.

Our first choice was to evaluate the Mean Squared Error (MSE) with respect to the previous iteration. We take the source point cloud and apply, separately, the rototranslations estimated during the current iteration of the algorithm and during the previous one. Therefore, we have the same point cloud in two different poses. Since applying a rototranslation, that is, a matrix multiplication, maintains the order of the points, we know that point $x^t_i$ in $X^t$ (the source point cloud aligned with the current estimate) corresponds to point $x^{t-1}_i$ in $X^{t-1}$ (the source point cloud aligned with the previous estimate). Hence, the point correspondences are known and exact. We used \cref{eq:msetermination}, where $N$ is the dimension of the point cloud, to calculate the Mean Squared Error (MSE) between two iterations.

\begin{equation}
\label{eq:msetermination}
MSE(X^t, X^{t-1}) = \frac{\sum_{i}^{N}||x_i^{t} - x_i^{t-1}||^2}{N}
\end{equation}

We stop the algorithm when the MSE drops under a certain relative threshold. With \emph{relative} we mean that we are not using a fixed absolute threshold, but we want to stop when, for example, the Mean Squared Error becomes smaller than a certain fraction of that at the previous iteration. 
That is:

\begin{equation}
    MSE(X^t, X^{t-1})\frac{<MSE(X^{t-1}, X^{t-2})}{threshold}
    \label{eq:msecondition}
\end{equation}

This means that we are stopping the algorithm when it is not able to move (or it is moving of a negligible amount) the source point cloud any more; thus, it has converged. We use a relative threshold, instead than an absolute, because it is much more flexible and does not have to be tuned for each set of point clouds.
However, instead than checking for \cref{eq:msecondition} just once, we ensure that the condition holds for several consecutive iterations. In this way we avoid stopping too early because of a single iteration during which the alignment was not improved, but that could be followed by other successful iterations.

Another option we evaluated is the use of the so-called \emph{Cost Drop}. During each outer iteration of the multi-iteration version of PPCR, an optimization problem is solved. Initially, the solution of the problem we are going to optimize will have a certain cost. The optimizer will, hopefully, reduce this cost to a lower value. The difference between the initial and the final cost is called \emph{Cost Drop}.
We used this value stopping the outer loop when the \emph{cost drop} of the inner loop drops under a threshold. We want to avoid absolute thresholds, since they need to be specifically tuned for each application. Instead, we express this threshold with respect to the initial cost of the problem: for example we could stop when the cost drop is less than $1\%$ of the initial cost of the problem. This is what we used for our experiments and proved to be a good threshold for obtaining accurate registrations. This condition is expressed by \cref{eq:costdrop}.
\begin{equation}
    \frac{cost(t)-cost(t-1)}{cost(t_0)} < {threshold}
    \label{eq:costdrop}
\end{equation}

Similarly to the MSE, and for the same reasons, this condition should hold for several iterations and not just once.

The third criterion we evaluated is the number of successful iterations of optimization problem. Solving an optimization problem with Levenberg-Marquard is an iterative process. Each step of this process can be successful, if the step managed to reduce the cost of the problem, or, otherwise, unsuccessful. We wanted to test if this value could be used as termination criterion somehow.

To evaluate the effectiveness of a termination criterion, we used the following idea. Suppose we have the ground truth for the source point cloud, \ie{}, we know the \emph{true} rototranslation between the reference frames of the source and target point clouds.
At the end of each iteration, we obtain an estimate of this rototranslation. Therefore, we can calculate the Mean Squared Error (as in \cref{eq:msecondition}) between our estimate and the ground truth, since they are the same point cloud in different poses.
Theoretically, if the algorithm is working properly, this error should decrease among the steps of the outer loop of the algorithm; therefore, the more the iterations, the smaller the difference becomes.
Practically, at some point this difference will cease to decrease, or, more precisely, it will start decreasing of a negligible amount. This is the iteration to which we should stop, since it means that the algorithm has converged to a solution. Note that it does not mean that it has converged to the right solution, but, nevertheless, that is the best solution we can get with the algorithm and the set of parameters we are using. 

Ideally, a good termination criterion should behave similarly to the difference \wrt{} the ground truth. It should stop the algorithm more or less at the same iteration that we would stop if we would be using the difference \wrt{} the ground truth (that, of course, in a real problem is unknown).

We evaluated the selected termination criteria on two datasets, to find which one works best. Eventually, we evaluated the best one on other datasets, to ensure that the results could be generalized and were not specific to the data we were using for the comparison and that the results we obtained were as good as if we were using a fixed high number of iterations.

\section{Results}

\begin{figure*}[!t]
	\centering
		\includegraphics[width=0.9\textwidth]{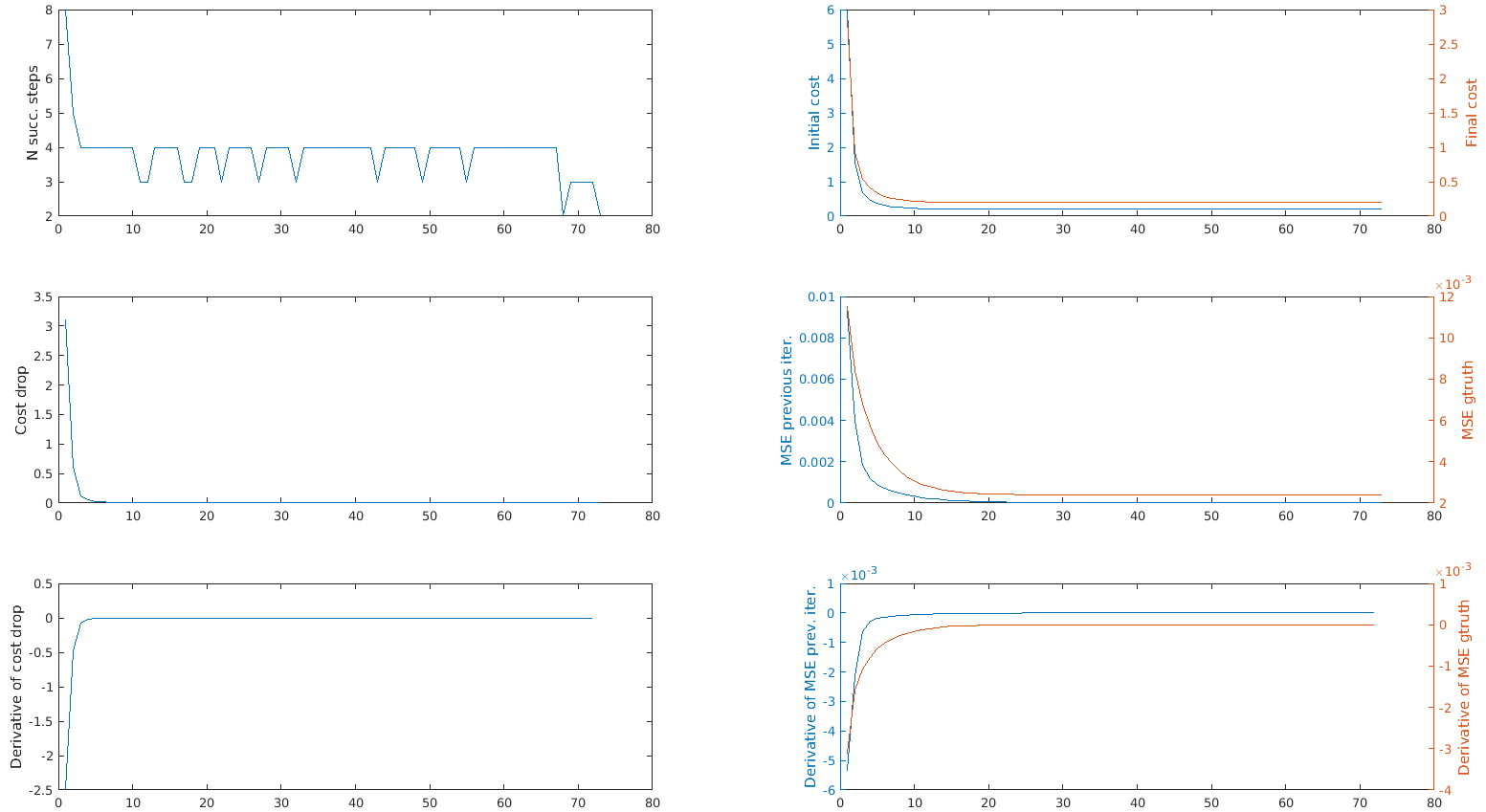}
	\caption{Plots of various termination criteria and their derivative for two point clouds from the Standford Bunny dataset.}
	\label{fig:bun045_termination}
\end{figure*}

In \cref{fig:bun045_termination} we plotted the three termination criteria while aligning two point clouds from the Standford Bunny dataset \cite{stanfordBunny}. The starting transformation between the two clouds is a rotation of $45\degree$ around the vertical axis.  
On the x-axis we have the number of the iteration, while on the y-axis we can find: the number of successful steps of the "inner" optimization problem, the initial and final cost of the "inner" optimization problem, the cost drop (\ie{}, the difference between the two previous values), the Mean Squared Error \wrt{} the previous iteration, the Mean Squared Error \wrt{} the ground truth and the discrete derivatives of the last three variables. We plotted also the discrete derivatives because they clearly show when a variable is not changing anymore: when the derivative becomes zero, the value of a variable has stabilized.

\begin{figure}
	\centering
		\includegraphics[width=0.2\textwidth]{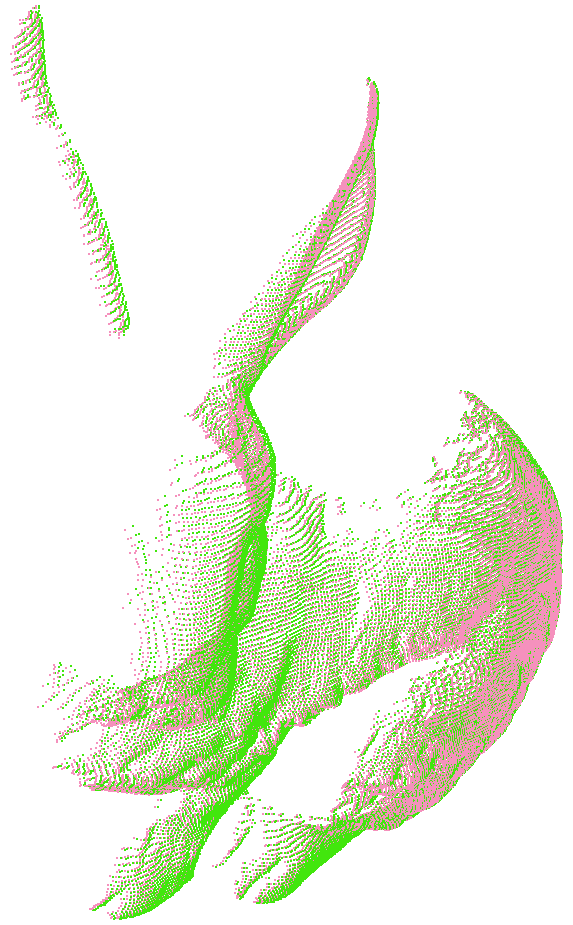}
	\caption{The same point cloud aligned with two different termination criteria: a large number of iterations (green point cloud) and our termination criterion based on the cost drop (pink point cloud).}
	\label{fig:bun045_aligned}
\end{figure}

We can see that both the cost drop and the MSE \wrt{} the previous iteration have a very similar trend to the MSE \wrt{} the ground truth. Most important, the three values stabilizes more or less at the same iteration. This is particularly obvious if we compare the discrete derivatives: they become almost zero more or less at the same time. Although the MSE \wrt{} to the ground truth keeps decreasing for a few iterations after the other two values stabilizes, its effect on the quality of the result is negligible. This becomes obvious looking at \cref{fig:bun045_aligned}, where we have two point clouds, one aligned using a predefined very large number of iterations, the second one using as stopping criterion the cost drop. We can seen that they overlap practically perfectly. The difference between the errors with respect to the ground truth of the two alignments is less than one tenth of the resolution of the point clouds, thus can be considered definitely negligible.
Other experiments on the same datasets yielded similar results.

Instead, the number of successful steps oscillates a lot and appears to be not correlated to the MSE \wrt{} the ground truth. For these reasons it was discarded.

\begin{figure}
	\centering
	\subfloat[]{
	\includegraphics[width=0.4\linewidth]{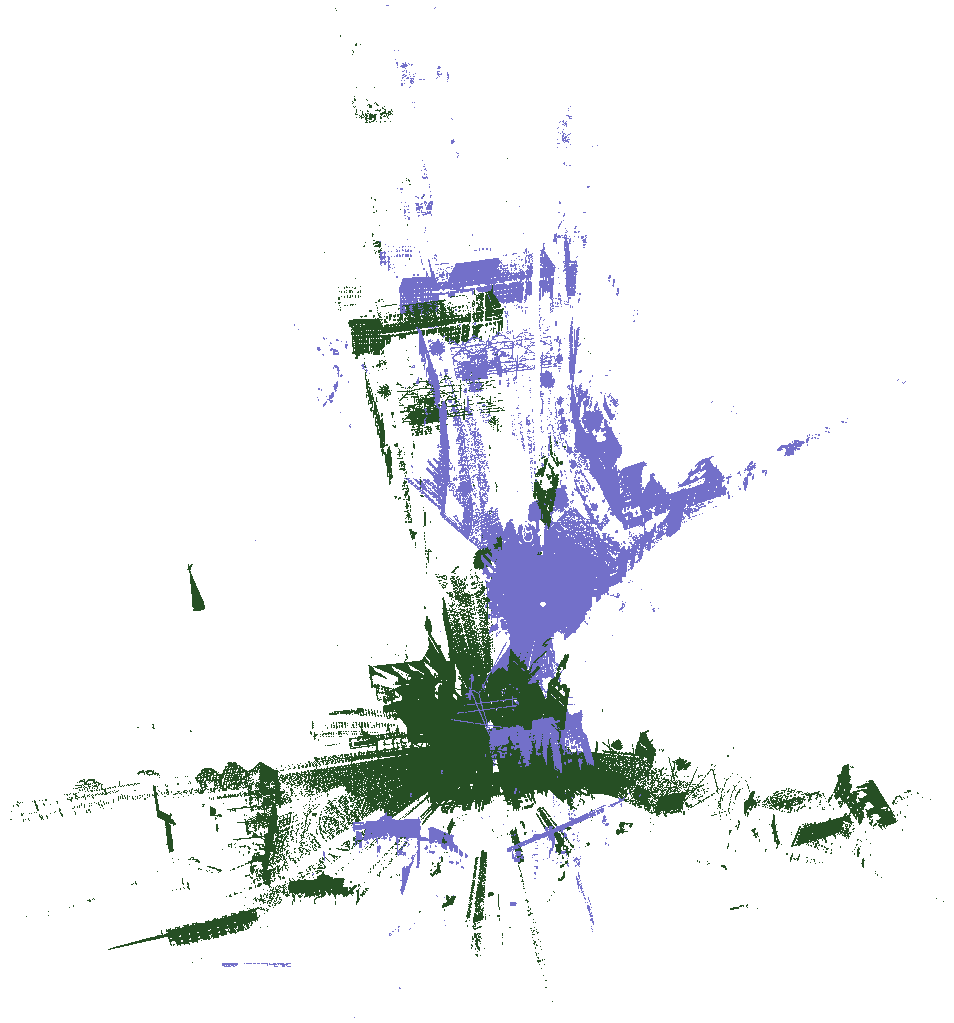}
	\label{fig:bremen_t}}
	\hfil
	\subfloat[]{
	\includegraphics[width=0.4\linewidth]{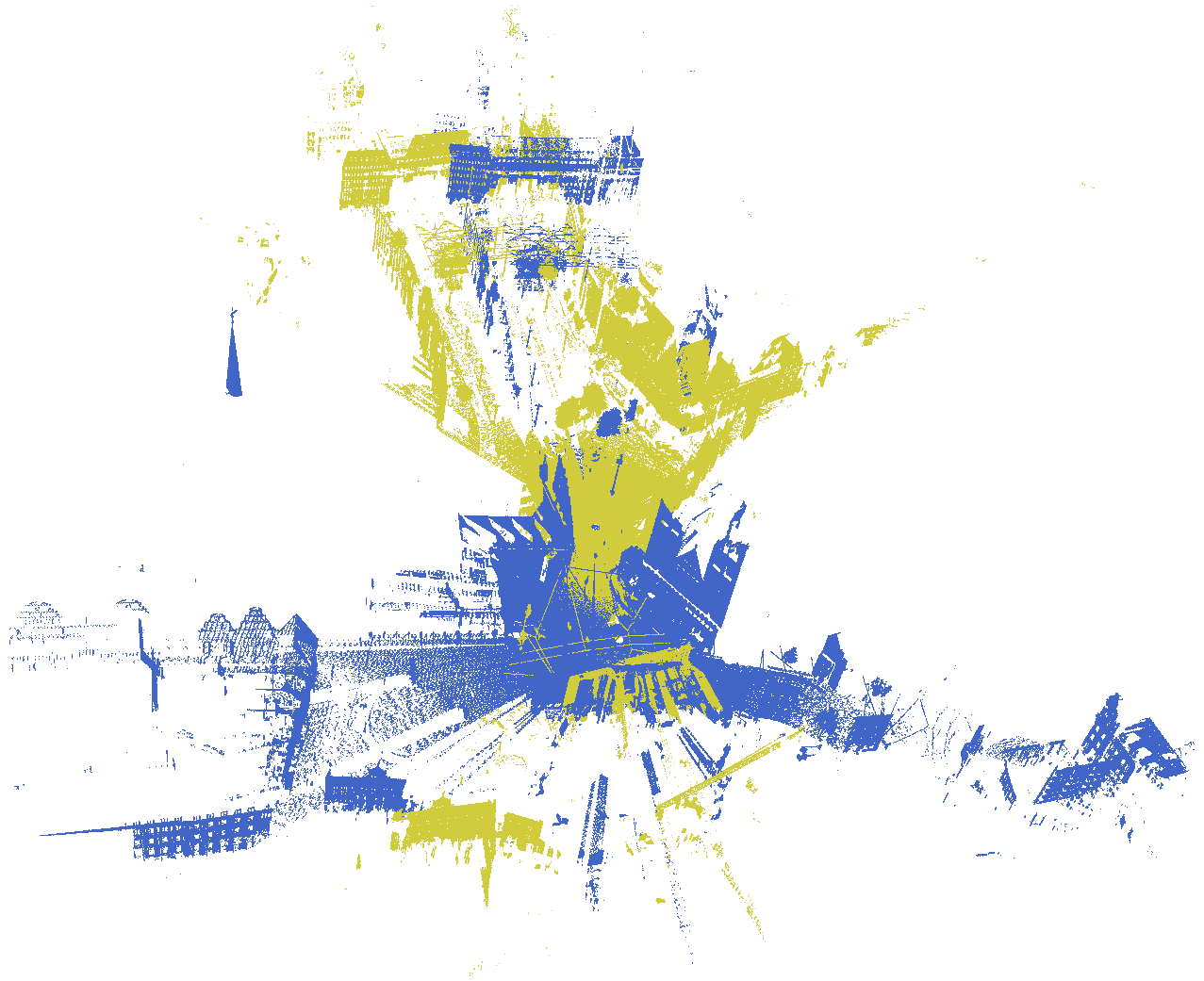}	
	\label{fig:bremen_r}}
	\caption{Two point clouds from the Bremen Dataset, to which we applied (a) a small translation (b) a small rotation.}
\end{figure}

\begin{figure}[]
	\centering
	\subfloat[]{
		\includegraphics[width=0.45\linewidth]{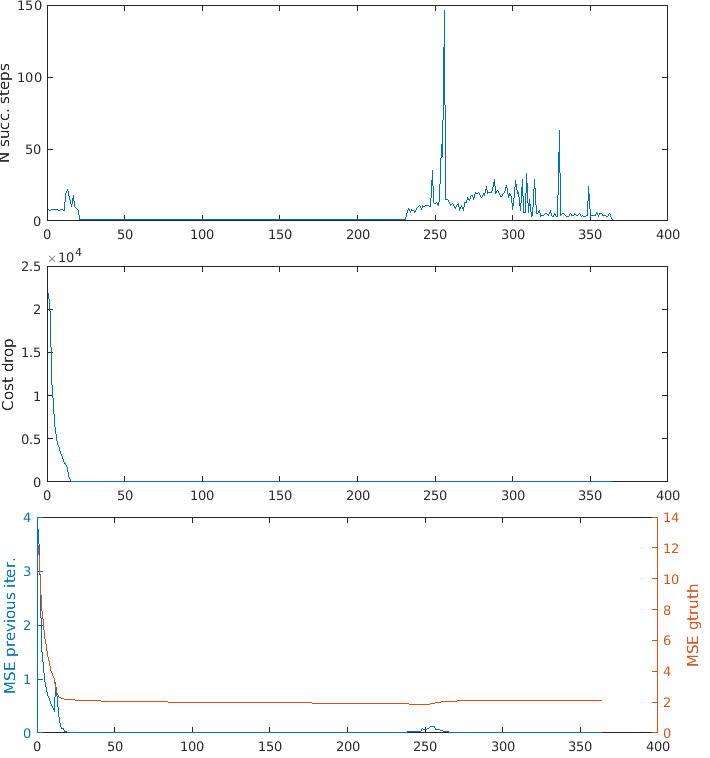}
	\label{fig:bremen_rot_term}
	}
	\hfil
	\subfloat[]{
	\includegraphics[width=0.45\linewidth]{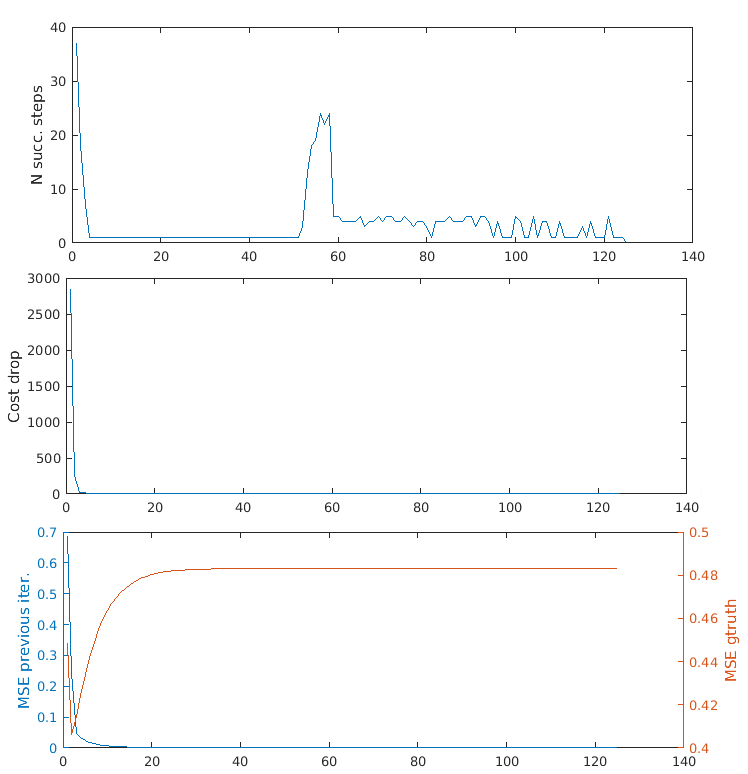}
	\label{fig:bremen_trans_term}
	}
	\caption{Termination criteria for the Bremen Dataset with (a) a small rotation (b) a small translation applied.}
\end{figure}

In \cref{fig:bremen_rot_term,fig:bremen_trans_term} we show the results obtained using the Bremen Dataset \cite{bremen}, to which we applied, respectively, a small rotation (\cref{fig:bremen_r}) and a small translation (\cref{fig:bremen_t}). In these plots, and in the followings, we will not show the derivatives for space reasons. We can see that the cost drop stabilizes more or less when also the MSE \wrt{} the previous iteration stabilizes, that is, when the cloud has already been moved to the right solution (future adjustments are negligible compared to the resolution of the point cloud).


Considering the results, there seems to be no strong reason to choose the MSE \wrt{} the previous iteration over the cost drop as termination criterion. However, it has to be considered that the MSE has to be specifically calculated after each iteration and is relatively computationally intensive, since the whole source point clouds has to be traversed.
This is not a computationally expensive operation \emph{per sé}, but, on the other hand, the relative cost drop is very fast to compute. Indeed, while solving an optimization problem we already calculate the absolute cost drop, since it is used as termination criterion of the inner loop by the optimization algorithm. Thus, calculating the relative cost drop requires only few more operations: it comes practically for free.
For this reason we have chosen to use the cost drop as termination criterion: it is very fast to compute and is as good as the Mean Squared Error.

We performed experiments also with clouds that the PPCR algorithm was not able to align properly. The reason is that we wanted to discover whether the termination criteria were able to stop the algorithm early enough, so that computational time is not wasted.

\begin{figure}
	\centering
	\subfloat[]{
	\includegraphics[width=0.4\linewidth]{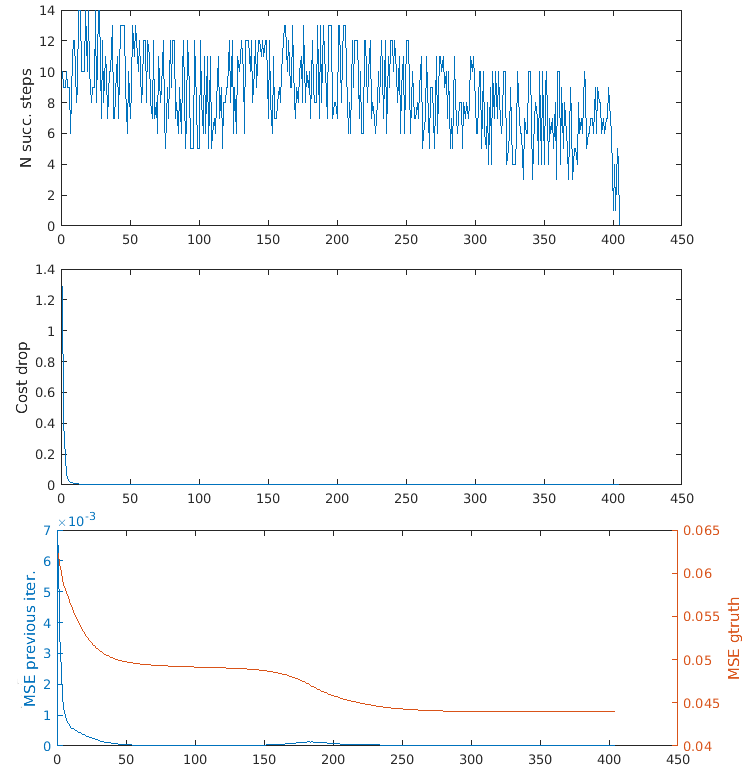}
	\label{fig:bun090_term}
	}
	\hfil
	\subfloat[]{
\includegraphics[width=0.4\linewidth]{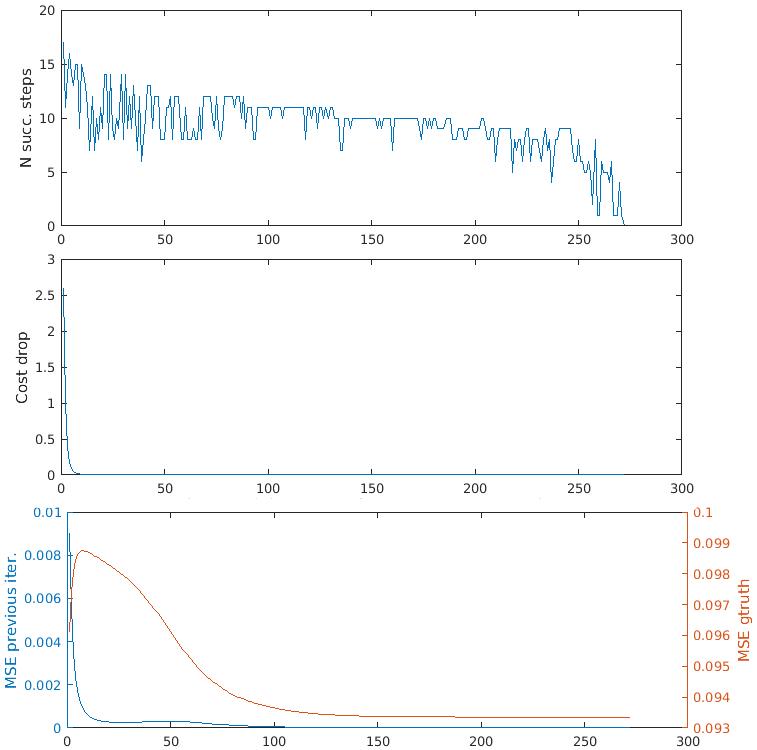}    \label{fig:bun180_term}}
	\caption{Termination criteria for two point clouds from the Standford Bunny Dataset. Initial misalignment of (a) $90\degree$ (b) $180\degree$.}
\end{figure}

As an example, we show the results on two point clouds from the Standford Bunny dataset, whose initial misalignment is a rotation of $90\degree$ around the vertical axis, (\cref{fig:bun090_term}), and a rotation of $180\degree$ around the vertical axis (\cref{fig:bun180_term}). In these cases, it can be seen that the cost drop stabilizes much earlier than the MSE \wrt{} the ground truth. This behaviour, indeed, is good, since it appeared only in unsuccessful alignments, during which stopping earlier is an advantage (going further would be only a waste of computational time).

We tested the chosen termination criterion on the same datasets we presented in our previous work \cite{agamennoni2016point} and on pairs of point clouds taken from the Stanford Bunny and Bremen datasets. Our goal is to show that the criterion is effective at stopping the algorithm at the right iteration: too late is a waste of computational time, too early leads to sub-optimal results. For this reason, we did not fine tune other parameters, since the performances of the algorithm were already shown in our previous work.

To show the effectiveness of our termination criteria, we executed the algorithm twice on each dataset. The first time using a predefined very large number of iterations. The second one using the cost drop to stop the algorithm. As a measure of the quality of the results we used the MSE \wrt{} the ground truth. The results are shown in \cref{tab:results_term,tab:results_fixed}.
\begin{table}[!t]
	\caption{Experimental results with termination criterion\label{tab:results_term}}
	\centering
	\begin{tabular}{| l || l | l |}
		\hline
		\textbf{Dataset} & \textbf{MSE} & \textbf{Iterations}\\
		\hline \hline
		Corridor & 0.31 & 14\\
		Office & 0.42 & 19\\
		Linkoping & 0.53 & 8\\
		Bunny & 0.0025 & 19\\
		Bremen & 0.45 & 8\\ \hline
	\end{tabular}
\end{table}
\begin{table}[!t]
	\caption{Experimental results with fixed number of iterations\label{tab:results_fixed}}
	\centering
	\begin{tabular}{| l || l | l |}
		\hline
		\textbf{Dataset} & \textbf{MSE} & \textbf{Iterations}\\
		\hline \hline
		Corridor & 0.66 & 100\\
		Office & 0.48 & 100\\
		Linkoping & 0.50 & 100\\
		Bunny & 0.0024 & 100\\
		Bremen & 0.48 & 100\\ \hline
	\end{tabular}
\end{table}
As it can be seen, the results using our criterion are usually comparable, and sometimes better, than when using many more iterations. This means that it succeeds at stopping the algorithm at the right iteration. In some cases, such as for the corridor dataset, the results in \cref{tab:results_term} are much better than those in table \cref{tab:results_fixed}. This happens because, sometimes, an excessive number of iterations is not only a waste of time, but could also bring the algorithm to converge to a wrong solution, even though the right solution was reached. This could happen with every algorithm that uses closest point based associations as a greedy approximation for the (unknown) correspondences.

\begin{figure}
	\centering
		\includegraphics[width=\linewidth]{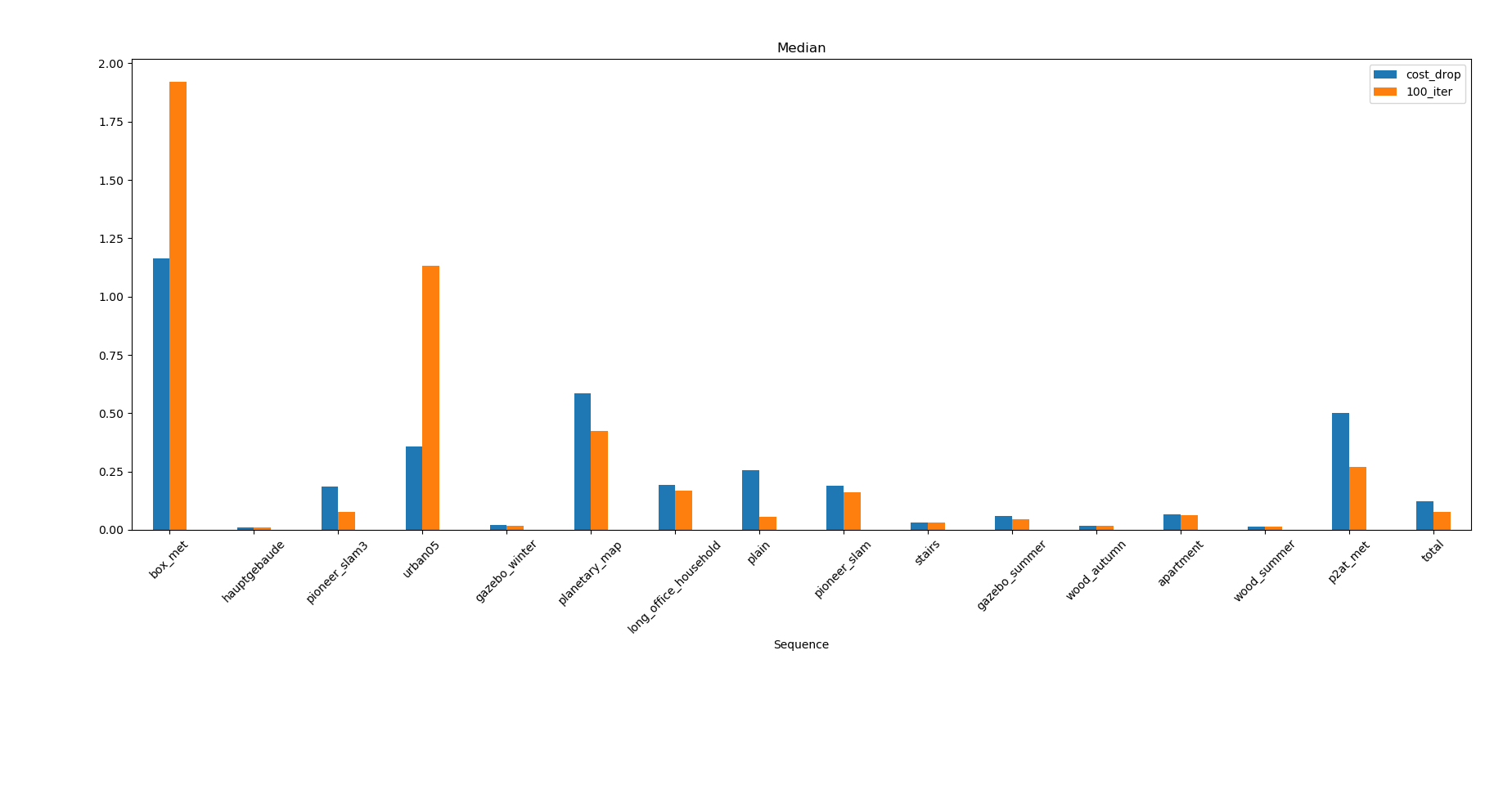}
	\caption{Histograms of the median results of PPCR on a comprehensive benchmark \cite{fontana2020benchmark} with two different termination criterion (100 iterations and the cost drop)}
	\label{fig:cost_drop_results}
\end{figure}

We performed experiments also on the Comprehensive Benchmark for Point Clouds Registration algorithms \cite{fontana2020benchmark}: it is composed of several point clouds, produced with different kinds of sensor and in different environments. Moreover, it includes several registration problems, with different initial misalignments and different overlaps between the clouds to align. For these reasons, we think that is particularly suitable to prove that the chosen criterion is at least as good as using a high number of iterations, but more efficient. Since the benchmark is composed of several datasets, we show statistics for both the single datasets and for the whole benchmark. The result is expressed in terms of median and $0.75$ and $0.95$ quantiles of the scaled mean squared error, as described in \cite{fontana2020benchmark}. In \cref{tab:median_result_benchmark} we compare the median of the results on the various datasets of the benchmark, using PPCR with a high fixed number of iterations ($100$ iterations) and using the cost drop as termination criterion (stopping when the cost drop is less the $1\%$ of the initial cost for more than $10$ iterations). For the cost drop, in the column named \emph{Number of iterations}, we show the mean number of iterations required to solve the registration problems. The same results are shown in \cref{fig:cost_drop_results} as histograms. For most sequences, the differences between the results obtained using the two methods are negligible. Indeed, the medians among all the registration problems of all the sequences (that is, the row named \emph{\textbf{total}} in \cref{tab:median_result_benchmark}) are very close. 

However, there are notable exceptions. For the \emph{box\_met} and the \emph{urban05} sequences, the cost drop leads to much better results (a lower median means a better alignment). This is the same behaviour we observed for the \emph{Corridor} dataset in the previous set of experiments. On the other hand, on the \emph{p2at\_met} and the \emph{plain} sequences the high number of iterations leads to better results. Nevertheless, it has to be considered that, even when the cost drop is not the best termination criterion, its results are still very good. At the same time, the average number of iterations required using the cost drop is $18.55$ and, considering the sequences individually, never greater than $31$; therefore there is a great reduction in computational time, \wrt{} using $100$ iterations.

\cref{tab:quant_result_benchmark} shows that, using the cost drop, we obtain even more consistent results, since its $0.95$ quantile is less than that of the $100$ iterations. The $0.75$ quantiles, instead, are very similar.

The proposed termination criterion requires two parameters: the percentage of drop and the number of iterations during which the condition described by \cref{eq:costdrop} should hold. However, the experiments show that using $1\%$ and $10$ iterations as thresholds leads to good results in a very large and varied set of registration problems. Therefore, this values should be adequate for most cases and should not require any further fine-tuning.

In \cref{tab:cost_drop_20} we show the results using $1\%$ and $20$ iterations as thresholds. The median result is very close to that obtained using $100$ iterations, although the mean number of iterations used is less than $30$; therefore, there is a great saving in computational time.  However, in our opinion, the difference \wrt{} using $10$ iterations as threshold is so negligible that it is not worth the extra computational time. Anyway, it is still an option if a very accurate result is desired.

PPCR using the proposed termination criterion, along with instructions on how to use it, is released on GitHub: \url{https://github.com/iralabdisco/probabilistic_point_clouds_registration}.

\begin{table}[]
    \centering
\begin{tabular}{|l|r|r|r|}
\hline
\textbf{Sequence}                         & \textbf{Cost drop} & \textbf{\begin{tabular}[c]{@{}r@{}}Number of\\ iterations\end{tabular}} & \textbf{100 iterations} \\ \hline\hline
\textbf{box\_met}                & 1.17               & 29.33                                                                   & 1.92                    \\
\textbf{hauptgebaude}            & 0.01               & 31.46                                                                   & 0.01                    \\
\textbf{pioneer\_slam3}          & 0.19               & 24.29                                                                   & 0.08                    \\
\textbf{urban05}                 & 0.36               & 21.81                                                                   & 1.13                    \\
\textbf{gazebo\_winter}          & 0.02               & 29.97                                                                   & 0.02                    \\
\textbf{planetary\_map}          & 0.59               & 14.82                                                                   & 0.42                    \\
\textbf{long\_office\_household} & 0.19               & 23.96                                                                   & 0.17                    \\
\textbf{plain}                   & 0.26               & 19.32                                                                   & 0.06                    \\
\textbf{pioneer\_slam}           & 0.19               & 27.27                                                                   & 0.16                    \\
\textbf{stairs}                  & 0.03               & 24.02                                                                   & 0.03                    \\
\textbf{gazebo\_summer}          & 0.06               & 25.11                                                                   & 0.04                    \\
\textbf{wood\_autumn}            & 0.02               & 29.34                                                                   & 0.02                    \\
\textbf{apartment}               & 0.07               & 23.04                                                                   & 0.06                    \\
\textbf{wood\_summer}            & 0.02               & 30.71                                                                   & 0.01                    \\
\textbf{p2at\_met}               & 0.50               & 18.55                                                                   & 0.27                    \\ \hline \hline
\textbf{total}                   & \textbf{0.12}      & \textbf{18.55}                                                          & \textbf{0.08}           \\\hline
\end{tabular}

    \caption{The median of the results of PPCR on a comprehensive benchmark \cite{fontana2020benchmark} with two different termination criterion (100 iterations and the cost drop)}
    \label{tab:median_result_benchmark}
\end{table}

\begin{table}[]
\centering
\begin{tabular}{|l|r|r|r|r|}
\hline
\textbf{Sequence}                & \textbf{\begin{tabular}[c]{@{}r@{}}0.75 quantile\\ (cost drop)\end{tabular}} & \textbf{\begin{tabular}[c]{@{}r@{}}0.95 quantile\\ (cost drop)\end{tabular}} & \textbf{\begin{tabular}[c]{@{}r@{}}0.75 quantile\\ (100 iterations)\end{tabular}} & \textbf{\begin{tabular}[c]{@{}r@{}}0.95 quantile\\ (100 iterations)\end{tabular}} \\ \hline \hline
\textbf{box\_met}                & 2.26                                                                         & 3.95                                                                         & 3.36                                                                              & 5.02                                                                              \\
\textbf{hauptgebaude}            & 0.03                                                                         & 0.72                                                                         & 0.02                                                                              & 0.78                                                                              \\
\textbf{pioneer\_slam3}          & 0.38                                                                         & 0.77                                                                         & 0.18                                                                              & 0.78                                                                              \\
\textbf{urban05}                 & 0.50                                                                         & 2.12                                                                         & 1.77                                                                              & 3.33                                                                              \\
\textbf{gazebo\_winter}          & 0.03                                                                         & 0.23                                                                         & 0.03                                                                              & 0.05                                                                              \\
\textbf{planetary\_map}          & 1.16                                                                         & 2.18                                                                         & 0.83                                                                              & 1.81                                                                              \\
\textbf{long\_office\_household} & 0.66                                                                         & 2.00                                                                         & 0.62                                                                              & 2.07                                                                              \\
\textbf{plain}                   & 0.50                                                                         & 0.94                                                                         & 0.20                                                                              & 1.00                                                                              \\
\textbf{pioneer\_slam}           & 0.43                                                                         & 3.54                                                                         & 0.45                                                                              & 4.68                                                                              \\
\textbf{stairs}                  & 0.09                                                                         & 0.24                                                                         & 0.09                                                                              & 0.24                                                                              \\
\textbf{gazebo\_summer}          & 0.20                                                                         & 0.65                                                                         & 0.13                                                                              & 1.02                                                                              \\
\textbf{wood\_autumn}            & 0.03                                                                         & 0.27                                                                         & 0.03                                                                              & 0.04                                                                              \\
\textbf{apartment}               & 0.29                                                                         & 1.30                                                                         & 0.27                                                                              & 2.02                                                                              \\
\textbf{wood\_summer}            & 0.02                                                                         & 0.27                                                                         & 0.02                                                                              & 0.03                                                                              \\
\textbf{p2at\_met}               & 1.04                                                                         & 2.00                                                                         & 0.84                                                                              & 2.29                                                                              \\ \hline \hline
\textbf{total}                   & \textbf{0.44}                                                                & \textbf{1.79}                                                                & \textbf{0.47}                                                                     & \textbf{2.38}                                                                     \\ \hline
\end{tabular}
\caption{The $0.75$ and $0.95$ quantiles of the results of PPCR on a comprehensive benchmark \cite{fontana2020benchmark} with two different termination criterion (100 iterations and the cost drop)}
    \label{tab:quant_result_benchmark}
\end{table}

\begin{table}[]
\centering
\begin{tabular}{|l|l|l|l|l|}
\hline
\textbf{Sequence}                & \textbf{median} & \textbf{0.75 quantile} & \textbf{0.95 quantile} & \textbf{iterations} \\ \hline \hline
\textbf{box\_met}                & 1.28            & 2.49                   & 4.27                   & 39.40               \\
\textbf{hauptgebaude}            & 0.01            & 0.02                   & 0.77                   & 42.35               \\
\textbf{pioneer\_slam3}          & 0.15            & 0.34                   & 0.75                   & 36.34               \\
\textbf{urban05}                 & 0.44            & 0.65                   & 4.31                   & 39.65               \\
\textbf{gazebo\_winter}          & 0.02            & 0.03                   & 0.05                   & 41.09               \\
\textbf{planetary\_map}          & 0.54            & 1.08                   & 2.07                   & 24.89               \\
\textbf{long\_office\_household} & 0.17            & 0.63                   & 1.99                   & 35.20               \\
\textbf{plain}                   & 0.17            & 0.46                   & 0.93                   & 31.78               \\
\textbf{pioneer\_slam}           & 0.18            & 0.44                   & 3.64                   & 38.13               \\
\textbf{stairs}                  & 0.03            & 0.09                   & 0.23                   & 34.08               \\
\textbf{gazebo\_summer}          & 0.05            & 0.19                   & 0.74                   & 35.96               \\
\textbf{wood\_autumn}            & 0.02            & 0.03                   & 0.06                   & 40.41               \\
\textbf{apartment}               & 0.06            & 0.28                   & 1.50                   & 33.36               \\
\textbf{wood\_summer}            & 0.01            & 0.02                   & 0.03                   & 42.44               \\
\textbf{p2at\_met}               & 0.47            & 0.97                   & 1.96                   & 29.75               \\ \hline \hline
\textbf{total}                   & 0.10            & 0.45                   & 1.86                   & 29.75               \\ \hline 
\end{tabular}
\caption{Results of PPCR on a comprenhnsive benchmark \cite{fontana2020benchmark}, using the cost drop with $1\%$ and $20$ iterations as thresholds}
\label{tab:cost_drop_20}
\end{table}

\section{Conclusions}
We introduced the use of the relative cost drop as termination criterion for the Probabilistic Point Clouds Registration Algorithm. We tested this criterion on different datasets and on a comprehensive benchmark for point clouds registration algorithms \cite{fontana2020benchmark}, which is composed of several registration problems, with different degrees of overlap and initial misalignment. The experiments prove that the cost drop is effective at stopping the algorithm at the right iteration, that is, when the algorithm has converged to a good solution that cannot be improved substantially anymore. Moreover, it stops the algorithm very early when solving problems that are not going to converge using more iterations, which is a very desirable behaviour. While it requires two parameters, we propose values that were proven to be effective on a wide range of registration problems.

\vspace{6pt} 



\authorcontributions{conceptualization, methodology, software and writing--original draft preparation S.F; writing--review, editing, supervision and project administration S.F., D.G.S..}

\funding{This research received no external funding.}


\conflictsofinterest{The authors declare no conflict of interest.} 

\abbreviations{The following abbreviations are used in this manuscript:\\

\noindent 
\begin{tabular}{@{}ll}
PPCR & Probabilistic Point Clouds Registration\\
ICP & Iterative Closest Point\\
MSE & Mean Squared Error
\end{tabular}}




\reftitle{References}


\externalbibliography{yes}
\bibliography{biblio}

\begin{thebibliography}{-------}
\providecommand{\natexlab}[1]{#1}

\bibitem[Besl and McKay(1992)]{besl1992method}
Besl, P.; McKay, N.D.
\newblock A method for registration of 3-D shapes.
\newblock {\em Pattern Analysis and Machine Intelligence, IEEE Transactions on}
  {\bf 1992}, {\em 14},~239--256.
\newblock
  doi:{\changeurlcolor{black}\href{https://doi.org/10.1109/34.121791}{\detokenize{10.1109/34.121791}}}.

\bibitem[Chen and Medioni(1991)]{chen1991object}
Chen, Y.; Medioni, G.
\newblock Object modeling by registration of multiple range images.
\newblock  Robotics and Automation, 1991. Proceedings., 1991 IEEE International
  Conference on. IEEE,  1991, pp. 2724--2729.

\bibitem[Zhang(1994)]{zhang1994iterative}
Zhang, Z.
\newblock Iterative point matching for registration of free-form curves and
  surfaces.
\newblock {\em International journal of computer vision} {\bf 1994}, {\em
  13},~119--152.

\bibitem[Agamennoni \em{et~al.}(2016)Agamennoni, Fontana, Siegwart, and
  Sorrenti]{agamennoni2016point}
Agamennoni, G.; Fontana, S.; Siegwart, R.Y.; Sorrenti, D.G.
\newblock Point clouds registration with probabilistic data association.
\newblock  Intelligent Robots and Systems (IROS), 2016 IEEE/RSJ International
  Conference on. IEEE,  2016, pp. 4092--4098.

\bibitem[Pomerleau \em{et~al.}(2013)Pomerleau, Colas, Siegwart, and
  Magnenat]{pomerleau_comparing_2013}
Pomerleau, F.; Colas, F.; Siegwart, R.; Magnenat, S.
\newblock Comparing {ICP} variants on real-world data sets.
\newblock {\em Autonomous Robots} {\bf 2013}, {\em 34},~133--148.
\newblock
  doi:{\changeurlcolor{black}\href{https://doi.org/10.1007/s10514-013-9327-2}{\detokenize{10.1007/s10514-013-9327-2}}}.

\bibitem[Segal \em{et~al.}(2009)Segal, Haehnel, and
  Thrun]{segal_generalized-icp._2009}
Segal, A.; Haehnel, D.; Thrun, S.
\newblock Generalized-{ICP}.
\newblock  Robotics: {Science} and {Systems},  2009, Vol.~2.

\bibitem[Biber and Stra{\ss}er(2003)]{biber2003normal}
Biber, P.; Stra{\ss}er, W.
\newblock The normal distributions transform: A new approach to laser scan
  matching.
\newblock  Intelligent Robots and Systems, 2003.(IROS 2003). Proceedings. 2003
  IEEE/RSJ International Conference on. IEEE,  2003, Vol.~3, pp. 2743--2748.

\bibitem[Merten(2008)]{merten2008three}
Merten, H.
\newblock The Three-Dimensional Normal-Distributions Transform.
\newblock {\em threshold} {\bf 2008}, {\em 10},~3.

\bibitem[Rusu \em{et~al.}(2008)Rusu, Blodow, Marton, and
  Beetz]{rusu_aligning_2008}
Rusu, R.; Blodow, N.; Marton, Z.; Beetz, M.
\newblock Aligning point cloud views using persistent feature histograms.
\newblock  {IEEE}/{RSJ} {International} {Conference} on {Intelligent} {Robots}
  and {Systems}, 2008. {IROS} 2008,  2008, pp. 3384--3391.
\newblock
  doi:{\changeurlcolor{black}\href{https://doi.org/10.1109/IROS.2008.4650967}{\detokenize{10.1109/IROS.2008.4650967}}}.

\bibitem[Rusu \em{et~al.}(2009)Rusu, Blodow, and Beetz]{rusu_fast_2009}
Rusu, R.; Blodow, N.; Beetz, M.
\newblock Fast {Point} {Feature} {Histograms} ({FPFH}) for 3D registration.
\newblock  {IEEE} {International} {Conference} on {Robotics} and {Automation},
  2009. {ICRA} '09,  2009, pp. 3212--3217.
\newblock
  doi:{\changeurlcolor{black}\href{https://doi.org/10.1109/ROBOT.2009.5152473}{\detokenize{10.1109/ROBOT.2009.5152473}}}.

\bibitem[Jiang \em{et~al.}(2009)Jiang, Cheng, and
  Chen]{jiang_registration_2009}
Jiang, J.; Cheng, J.; Chen, X.
\newblock Registration for 3-{D} point cloud using angular-invariant feature.
\newblock {\em Neurocomputing} {\bf 2009}, {\em 72},~3839--3844.
\newblock
  doi:{\changeurlcolor{black}\href{https://doi.org/10.1016/j.neucom.2009.05.013}{\detokenize{10.1016/j.neucom.2009.05.013}}}.

\bibitem[Fischler and Bolles(1981)]{fischler1981random}
Fischler, M.A.; Bolles, R.C.
\newblock Random sample consensus: a paradigm for model fitting with
  applications to image analysis and automated cartography.
\newblock {\em Communications of the ACM} {\bf 1981}, {\em 24},~381--395.

\bibitem[Zeng \em{et~al.}(2017)Zeng, Song, Nie{\ss}ner, Fisher, Xiao, and
  Funkhouser]{zeng20173dmatch}
Zeng, A.; Song, S.; Nie{\ss}ner, M.; Fisher, M.; Xiao, J.; Funkhouser, T.
\newblock 3dmatch: Learning local geometric descriptors from rgb-d
  reconstructions.
\newblock  Proceedings of the IEEE Conference on Computer Vision and Pattern
  Recognition,  2017, pp. 1802--1811.

\bibitem[Gojcic \em{et~al.}(2019)Gojcic, Zhou, Wegner, and
  Wieser]{gojcic2019perfect}
Gojcic, Z.; Zhou, C.; Wegner, J.D.; Wieser, A.
\newblock The perfect match: 3d point cloud matching with smoothed densities.
\newblock  Proceedings of the IEEE Conference on Computer Vision and Pattern
  Recognition,  2019, pp. 5545--5554.

\bibitem[Aoki \em{et~al.}(2019)Aoki, Goforth, Srivatsan, and
  Lucey]{aoki2019pointnetlk}
Aoki, Y.; Goforth, H.; Srivatsan, R.A.; Lucey, S.
\newblock Pointnetlk: Robust \& efficient point cloud registration using
  pointnet.
\newblock  Proceedings of the IEEE Conference on Computer Vision and Pattern
  Recognition,  2019, pp. 7163--7172.

\bibitem[Sarode \em{et~al.}(2019)Sarode, Li, Goforth, Aoki, Srivatsan, Lucey,
  and Choset]{sarode2019pcrnet}
Sarode, V.; Li, X.; Goforth, H.; Aoki, Y.; Srivatsan, R.A.; Lucey, S.; Choset,
  H.
\newblock Pcrnet: point cloud registration network using pointnet encoding.
\newblock {\em arXiv preprint arXiv:1908.07906} {\bf 2019}.

\bibitem[Holz \em{et~al.}(2015)Holz, Ichim, Tombari, Rusu, and
  Behnke]{holz2015registration}
Holz, D.; Ichim, A.E.; Tombari, F.; Rusu, R.B.; Behnke, S.
\newblock Registration with the point cloud library: A modular framework for
  aligning in 3-d.
\newblock {\em IEEE Robotics \& Automation Magazine} {\bf 2015}, {\em
  22},~110--124.

\bibitem[Zhou \em{et~al.}(2016)Zhou, Park, and Koltun]{zhou2016fast}
Zhou, Q.Y.; Park, J.; Koltun, V.
\newblock Fast global registration.
\newblock  European Conference on Computer Vision. Springer,  2016, pp.
  766--782.

\bibitem[Yang \em{et~al.}(2020)Yang, Shi, and Carlone]{Yang20arXiv-TEASER}
Yang, H.; Shi, J.; Carlone, L.
\newblock TEASER: Fast and Certifiable Point Cloud Registration {\bf 2020}.
\newblock  \href{http://xxx.lanl.gov/abs/2001.07715}{{\normalfont
  [arXiv:cs.RO/2001.07715]}}.

\bibitem[Turk()]{stanfordBunny}
Turk, G.
\newblock The Stanford Bunny.
\newblock \url{http://www.cc.gatech.edu/~turk/bunny/bunny.html}.

\bibitem[Borrmann and Neuchter()]{bremen}
Borrmann, D.; Neuchter, A.
\newblock The Bremen Dataset.
\newblock \url{http://kos.informatik.uni-osnabrueck.de/3Dscans/}.

\bibitem[Fontana \em{et~al.}(2020)Fontana, Cattaneo, Ballardini, Vaghi, and
  Sorrenti]{fontana2020benchmark}
Fontana, S.; Cattaneo, D.; Ballardini, A.L.; Vaghi, M.; Sorrenti, D.G.
\newblock A Benchmark for Point Clouds Registration Algorithms.
\newblock {\em arXiv preprint arXiv:2003.12841} {\bf 2020}.

\end{thebibliography}



\end{document}